\documentclass[
  journal=medium,
  year=2025,
]{cup-journal}

\makeatletter
\renewcommand*\cup@reference@code{
  \RequirePackage[style=authoryear,backend=biber,natbib]{biblatex}
  \renewcommand*{\bibfont}{\footnotesize}
}
\makeatother

\usepackage{amsmath}
\usepackage[nopatch]{microtype}
\usepackage{booktabs}

\title{Atomizer: Generalizing to new modalities by breaking satellite images down to a set of scalars}

\author{Hugo Riffaud de Turckheim}
\affiliation{INRIA, Montpellier, France}
\email[Hugo Riffaud de Turckheim]{hugo.riffaud--de-turckheim@inria.fr}

\author{Sylvain Lobry}
\affiliation{LIPADE, Paris, France}

\author{Roberto Interdonato}
\affiliation{CIRAD, Montpellier, France}

\author{Diego Marcos}
\affiliation{INRIA, Montpellier, France}

\begin{document}

\begin{abstract}
The growing number of Earth observation satellites has led to increasingly diverse remote sensing data, with varying spatial, spectral, and temporal configurations. Most existing models rely on fixed input formats and modality-specific encoders, which require retraining when new configurations are introduced, limiting their ability to generalize across modalities. We introduce Atomizer, a flexible architecture that represents remote sensing images as sets of scalars, each corresponding to a spectral band value of a pixel. Each scalar is enriched with contextual metadata (acquisition time, spatial resolution, wavelength, and bandwidth), producing an atomic representation that allows a single encoder to process arbitrary modalities without interpolation or resampling. Atomizer uses structured tokenization with Fourier features and non-uniform radial basis functions to encode content and context, and maps tokens into a latent space via cross-attention. Under modality-disjoint evaluations, Atomizer outperforms standard models and demonstrates robust performance across varying resolutions and spatial sizes.
\end{abstract}

\section{Introduction}

Earth observation satellites capture our planet through different lenses, each offering distinct spectral, spatial, and temporal configurations. Many satellites are being launched with diverse spectral, spatial and temporal resolutions. While the first Earth observation satellite has been launched in 1972, more than half of them have been launched since 2020 (with 105 planned launches in 2025)\cite{usgscomp}. Some satellites prioritize rich spectral information across dozens of bands but sacrifice spatial detail (e.g., MODIS), while others offer fine-grained spatial resolution with fewer spectral bands (e.g., Sentinel-2). These particularities are double-edged: they provide complementary insights for applications such as drought monitoring \cite{li2025improved}, where MODIS’s frequent coverage is combined with Sentinel-2’s spatial detail to detect field-scale crop stress, as well as plant species classification \cite{pazos2024planted} and crop mapping \cite{skakun2017combined}. Yet traditional vision models, designed for uniform input formats, struggle to accommodate this heterogeneity.
Recent work has begun to incorporate prior knowledge such as resolution or wavelengths into learning pipelines \cite{astruc2024anysat, reed2023scale}. These methods often embed spatial resolution and spectral information through specialized modules within their model architectures. While these methods are suitable for some aspects of heterogeneity, most of them still rely on modality-specific encoders and patch representations, limiting their flexibility to unseen sensors configurations, often requiring costly pre-training steps \cite{li2025fleximo}.

\noindent We propose a different solution: instead of proposing a specific module for each type of input modality, we adapt the representation itself. Atomizer introduces a unified token-based framework that decomposes each observation into its most elementary unit, the reflectance of the band of a given pixel. Each token is enriched with acquisition-specific metadata, including spatial resolution, central wavelength, and spectral bandwidth. This atomic representation preserves all contextual information while enabling a simple, consistent framework which can be applied to a new optical modality that has a different combination of resolution, image size and spectral bands than those seen at training time.
Atomizer builds on Perceiver \cite{jaegle2021perceiver}, taking the core idea, the breaking down of images to a set of pixels, to the extreme, by mapping the set of scalars that compose an image into a compact latent space via cross-attention. This enables the model to handle inputs of arbitrary spatial size and channel depth. Metadata is encoded using structured strategies: Fourier features \cite{tancik2020fourier} for positional encodings, and radial basis functions for spectral properties. This flexible framework allows each type of prior knowledge to be explicitly represented, making it straightforward to incorporate expert domain insights into the encoding process. As a result, Atomizer can reason across heterogeneous satellite observations without requiring architectural changes or retraining for new modalities.

\noindent To test Atomizer's generalization capabilities, we design experiments that simulate real-world scenarios where models must handle data from entirely different satellites than those used during training. Our modality-disjoint protocol ensures that training and test sets contain observations from distinct satellite configurations with different spatial dimensions, resolutions, and spectral band combinations. Under these conditions, conventional architectures show significant performance degradation, while Atomizer maintains consistent accuracy even when processing previously unseen modality combinations. This resilience to varying input characteristics demonstrates Atomizer's potential for operational Earth observation systems where sensor configurations continuously evolve.
Our main contributions are:
\begin{itemize}
\item We introduce Atomizer, a token-based architecture that represents remote sensing observations at their native scale and structure, eliminating the need for retraining as new satellite missions emerge.
\item We design a comprehensive metadata encoding scheme using Fourier features, positional encodings, and non-uniform RBFs to capture acquisition-specific context.
\item We establish a rigorous modality-disjoint evaluation protocol, demonstrating Atomizer’s ability to generalize across unseen sensor types and configurations.
\end{itemize}
\section{Related Work}

\textbf{Addressing Resolution Variability.} A limitation of most computer vision approaches for Earth observation, such as those based on convolutional neural networks (CNN) or Vision Transformers (ViT)~\cite{dosovitskiy2020image}, is their inability to effectively handle varying Ground Sample Distances (GSDs) at inference time. The fixed patch size selected during pre-training forces a trade-off. In high-resolution images, small patches lead to significant computational overhead, while in low-resolution, larger patches risk discarding critical fine-grained details. Several recent methods have attempted to address this resolution variability by incorporating GSD as prior knowledge. SenPa-MAE \cite{prexl2024senpa} incorporates resolution as an explicit feature vector generated through multi-layer perceptrons, which is then added to the patch embeddings. Other approaches encode resolution information directly within the positional embeddings associated with each patch; this approach is taken by ScaleMAE \cite{reed2023scale,tseng2025galileo,astruc2024anysat}, allowing these models to adapt feature extraction based on input resolution. FlexiMo \cite{li2025fleximo} takes a different approach by applying bilinear interpolation to both raw image patches and patch embedding weights, enabling dynamic adjustment to varying spatial resolutions. While these methods improve performance across different GSD configurations, they remain constrained by their reliance on rigid patch-based tokenization.

\noindent
\textbf{Incorporating Spectral Heterogeneity.} Unlike natural images, satellite data contains spectral information that varies significantly across sensors \cite{liu2015survey}, with different bands capturing distinct portions of the electromagnetic spectrum defined by their central wavelengths and bandwidths. Therefore, incorporating spectral configuration as prior knowledge is essential for effectively processing heterogeneous remote sensing data. Several approaches have emerged to address this challenge. FlexiMo \cite{li2025fleximo} incorporates spectral information by using central wavelength to generate convolution kernels dynamically, enabling adaptation to varying channel configurations. Galileo \cite{tseng2025galileo} splits input bands into semantically cohesive groups (e.g., RGB bands in Sentinel-2), enabling the network to model relationships between spectrally related channels more effectively. However, this approach introduces additional design complexity in determining optimal channel groupings and lacks flexibility when confronted with novel spectral configurations not encountered during training. SenPa-MAE \cite{prexl2024senpa} embeds spectral responses as feature vectors added to patch embeddings, akin to positional encodings. While these methods improve handling of spectral variability, they typically require complex architectural modifications and still struggle to process truly heterogeneous observations with arbitrary band combinations without retraining.

\noindent
\textbf{Handling Variable Spatial Dimensions.} Beyond resolution and spectral variations, remote sensing models must also handle input images of varying spatial dimensions. Traditional vision transformers require fixed-size inputs, necessitating resizing or cropping that can distort spatial relationships or lose information. Several approaches have been proposed to address this limitation. FlexiMo and Galileo \cite{li2025fleximo,tseng2025galileo} 
simply resize the patch embedding weights to fit with the dimensions of the input. AnySat \cite{astruc2024anysat} takes a different approach by using fixed patch sizes that represent consistent ground distances in meters rather than pixel counts, allowing the model to maintain physical scale awareness across different resolution images. As a more significant departure from patch-based paradigms, Presto \cite{tseng2023lightweight} eliminates patches entirely by encoding each pixel individually. While these approaches show strong results for handling varying input shapes, they typically require complex architectural modifications or introduce computational inefficiencies when processing high-resolution inputs.

\noindent
\textbf{Towards Modality-Agnostic Architectures.} While the approaches described above make progress in handling specific aspects of satellite data, most of them rely on complex modules added within the ViT framework, gradually increasing architectural complexity with each new modality characteristic. This approach contradicts the principle of Occam's razor that simpler solutions are preferable. Instead of further complicating existing architectures, we draw inspiration from Perceiver \cite{jaegle2021perceiver}, a model explicitly designed to process any modality: audio, video, images, through a unified framework. Perceiver addresses the computational scaling challenges of attention mechanisms by mapping arbitrary-sized inputs to a compact latent representation through cross-attention, enabling it to handle diverse input types without specialized encoders. By enriching inputs with modality-specific metadata through Fourier features \cite{tancik2020fourier} rather than architectural modifications, Perceiver maintains a consistent architecture across modalities. Our approach leverages this insight while introducing a tokenization scheme specifically designed for the unique challenges of remote sensing data.






\begin{figure}[ht]
  \centering
  \includegraphics[width=1.0\textwidth]{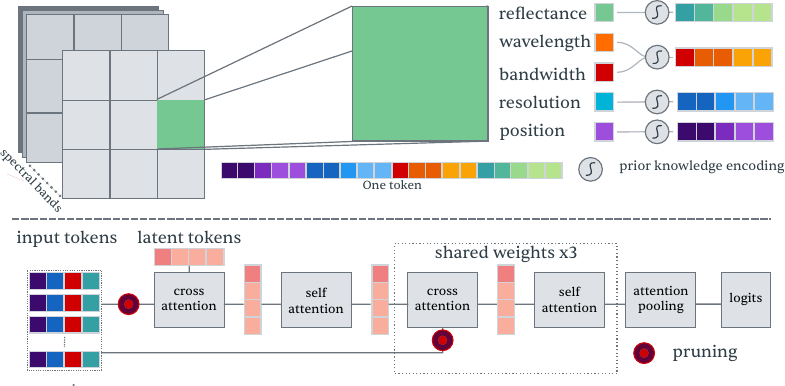}
  \caption{Top:  Atomizer's token-based representation. Each spectral band of each pixel is decomposed into a token that encodes multiple attributes: reflectance (band value), wavelength, bandwidth, resolution, and position. Bottom: Diagram of the used architecture.} 
  \label{fig:method_overview}
\end{figure}

\section{Methodology}

\noindent
Rather than developing increasingly complex architectural adaptations to accommodate heterogeneous satellite data, we propose a different approach that rethinks the input representation itself. Inspired by the Perceiver model \cite{jaegle2021perceiver}, we bypass the need for fixed input formats by representing each observation as a fine-grained token. Specifically, we construct a token for each band of each pixel. These tokens include not only the raw band value, but also metadata such as spatial resolution, wavelength, and bandwidth, capturing both the content and context of each observation.
To encode these attributes, we employ two different basis functions: (i) Fourier features for the band of value (e.g. reflectance) and the positional encoding; (ii) Radial basis functions for spectral properties such as wavelength and bandwidth.
Our token-based design decouples the model from constraints such as fixed input shapes or specific acquisition schedules. By feeding the model a flat list of tokens, we can process input images of arbitrary resolution, spatial extent and spectral composition. This enables training a single encoder across diverse remote sensing datasets without requiring resampling, interpolation, or modality-specific adaptations.

\subsection{Token Construction}

To represent the highly heterogeneous nature of remote sensing data, we build a token for every spectral band of every pixel. Each token is designed to capture not only the observed reflectance or radiance value, but also metadata that contextualizes the observation.\\

\noindent Formally, for a given image $I$ with spatial coordinates $(x, y)$ at resolution $r$ and spectral band $b$, we define the token $\mathbf{z}_{xyb}$ as:

\[
\mathbf{z}_{xyb} = \text{Concat} \left( \phi_I(I_{xyb}), \phi_\text{res}(r), \phi_\lambda(\lambda_b, \Delta\lambda_b) \right)
\]
Where $I_{xyb}$ is the raw band value, $\phi_\text{res}(r)$ encodes spatial resolution and the position of the band value within its original image, and $\phi_\lambda(\lambda_b, \Delta\lambda_b)$ encodes the spectral properties: central wavelength $\lambda_b$ and bandwidth $\Delta\lambda_b$.
This formulation allows us to make use of the metadata of each observation, transforming raw image bands into a structured, interpretable, and learnable representation.


\noindent
\textbf{Fourier Features.}
To encode continuous scalar metadata (such as time and resolution), we use \textit{Fourier features}, also known as sinusoidal position encodings \cite{tancik2020fourier}. These project a scalar input into a higher-dimensional space using sine and cosine functions, allowing the model to capture both low- and high-frequency variations.

\noindent Let $\tilde{x,} \in [-1, 1]$ be a normalized scalar,$L$ the number of frequency components and $f_{\text{max}}$ the maximum frequency component. The Fourier embedding function $\gamma(\cdot)$ is defined as:

\[
\gamma(\tilde{x};L,f_\text{max}) = \left[ \sin(\pi f_1 \tilde{x}), \cos(\pi f_1 \tilde{x}), \dots, \sin(\pi f_L \tilde{x}), \cos(\pi f_L \tilde{x}) \right]
\]
Where the $L$ frequencies $f_i$ are linearly spaced between 1 and $f_\text{max}$.






\noindent
\textbf{Reflectance encoding} $\phi_I$: the reflectance of the band $b$ at position $x,y$ within the image is encoded with using Fourier features, i.e. $\phi_I(I_{xyb})=\gamma({I_{xyb}};L,f_\text{max})$.

\noindent
\textbf{Spatial resolution encoding} $\phi_\text{res}$:
To encode spatial resolution and position information, we implement a resolution-aware positional encoding scheme that incorporates ground sampling distance (GSD) as an explicit factor in Fourier features.

\noindent For each pixel at coordinates $(x, y)$ with resolution $g$, we compute resolution-scaled coordinates that span the physical extent of the image:
$$
x_\text{scaled} = \left(x - \frac{w}{2}\right) \cdot \frac{g}{G}, \quad y_\text{scaled} = \left(y - \frac{h}{2}\right) \cdot \frac{g}{G}
$$
where $w$ and $h$ are the width and height of the image in pixels, $g$ is the ground sampling distance in meters per pixel, and $G$ is a reference normalization constant. The resolution-modulated Fourier features are then computed as:
$$
\phi_\text{res}(x,y,g) = \text{Concat}\left(\gamma(x_\text{scaled}; L, f_{\text{max}}), \gamma(y_\text{scaled}; L, f_{\text{max}})\right)
$$
where $\gamma(\cdot)$ is the Fourier encoding function, $L$ is the number of frequency components, and $f_{\text{max}}$ is the maximum frequency.\\

\noindent  The scaling factor $\frac{g}{G}$ directly modulates the coordinate space based on physical resolution. At finer resolutions (small $g$), coordinates span a smaller range corresponding to a smaller physical area, while at coarser resolutions (large $g$), coordinates span a larger range reflecting greater physical coverage. This enables the model to maintain consistent spatial understanding across different sensor configurations. The encoding yields a vector of size $4L+2$, where we concatenate the original pixel coordinates to preserve absolute positional information.\\

\noindent
\textbf{Spectral encoding} $\phi_\lambda$:  
To encode the spectral characteristics of each band, defined by its central wavelength $\lambda$ and bandwidth $\mu$, we use a radial basis function (RBF) encoding scheme based on Gaussian kernels. We use radial basis functions $\phi_\lambda(\lambda, \mu)$ to encode spectral characteristics. Each band's spectral support (defined by central wavelength $\lambda$ and bandwidth $\mu$) is matched against strategically placed Gaussians.\\

\[
\phi_\lambda(\lambda, \mu) = \left[\int_{\lambda-\mu/2}^{\lambda+\mu/2} \mathcal{N}(\mu_1, \sigma_1) d\lambda', \ldots, \int_{\lambda-\mu/2}^{\lambda+\mu/2} \mathcal{N}(\mu_k, \sigma_k) d\lambda'\right]
\]

\noindent Unlike typical uniform RBF placements, our Gaussians are strategically distributed: we allocate more narrow basis functions in spectral regions where many sensors and modalities operate (e.g., between 400\,nm and 800\,nm in the visible range). Fewer and wider Gaussians are used in sparsely populated spectral zones. This non-uniform distribution ensures higher resolution where it's most needed, enabling the model to distinguish between closely spaced bands. For each Gaussian $i$, we compute the integral of the Gaussian function over the band's spectral support range $[\lambda - \mu/2, \lambda + \mu/2]$. This yields a $k$-dimensional feature vector where each component reflects how much the band overlaps with a particular Gaussian. The final spectral encoding is then L2-normalized.\\

\noindent  By jointly encoding the bandwidth alongside the central wavelength, this representation allows the model to differentiate between bands with similar center wavelengths but different spectral widths. This is especially important in the visible spectrum, where sensors like Sentinel-2, MODIS, and Landsat 8 may use overlapping wavelengths but with significantly different bandwidths. This enables the model to better align and reason across sensors, even when they operate in overlapping spectral regions.

\subsection{Architecture}

\noindent The encoded tokens are processed using a Perceiver-style architecture \cite{jaegle2021perceiver}. Our model ingests the unordered set of tokens and maps them into a compact latent representation through cross-attention mechanisms. Specifically, a set of $L$ learnable latent tokens attends to the input tokens, capturing information from the entire token set regardless of its size or structure.


\noindent To generate prediction logits from this latent space, we employ attention pooling as described in \cite{chen2023self}, where the latent vectors are aggregated via attention mechanisms to produce classification outputs.

\noindent A challenge in our approach is the potentially large number of tokens generated for high-resolution or spectrally-rich images. To address this computational limitation, we implement a token pruning strategy during training. Before each cross-attention operation, we randomly remove a proportion $p$ of the input tokens. This masking is applied independently at each layer, forcing the model to learn robust representations despite seeing only a subset of tokens at each step. In our experiments, we set $p=0.5$, which substantially reduces memory requirements while maintaining performance. This pruning approach, combined with the Perceiver's latent bottleneck design and our weight sharing mechanism, enables efficient processing of inputs with arbitrary numbers of tokens without compromising the model's representational capacity.

\section{Experimental Framework}

\begin{figure}[ht]
  \centering
  \includegraphics[width=1.0\textwidth]{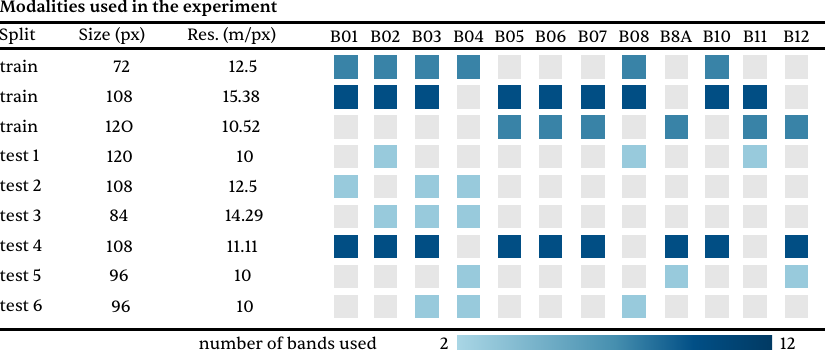}
  \caption{Modalities used for training and testing. Each row represents a distinct modality configuration with its spatial dimensions (in pixels), ground sampling distance (meters per pixels), and spectral bands. The shade of blue represents the total number of bands (2-12).}
  \label{fig:dataset_overview}
\end{figure}

To evaluate Atomizer's ability to generalize to unseen sensor configurations, we design a modality-disjoint evaluation protocol using the BigEarthNet dataset \cite{clasen2024reben}, a large-scale benchmark for remote sensing image classification containing Sentinel-2 multispectral imagery across 19 land cover classes. Our experimental setup comprises 69,373 images for training, 65,618 for validation, and 63,972 for testing, with the complete dataset and modality configurations to be made publicly available on GitHub.
We define distinct modalities by systematically varying three attributes: (1) spatial dimensions (pixel size), (2) ground sampling distance (GSD), and (3) spectral band composition (subset of Sentinel-2 bands). Each modality represents a unique combination of these attributes, simulating different satellite sensor configurations. This approach allows us to evaluate how models perform when encountering new sensor characteristics not seen during training.
Our experimental design ensures that modalities used for training and testing are completely disjoint. We define separate training and evaluation modalities, with each training image associated exclusively with one training modality. By ensuring no image appears under multiple modalities during training, we prevent the model from explicitly learning relationships between different modality configurations. Test images are associated with entirely unseen modalities, creating a true cross-modality generalization challenge. Figure \ref{fig:dataset_overview} illustrates the distinct modalities used for training and testing, highlighting their varying characteristics in terms of resolution, spatial dimensions, and spectral composition.

\noindent We evaluate performance on the multilabel classification task of BigEarthNet, reporting mean Average Precision (mAP) and overall accuracy metrics. For comparison, we implement three baseline architectures: (1) a small ResNet, representing CNN-based approaches; (2) a standard Vision Transformer (ViT), representing patch-based attention methods; and (3) ScaleMAE, representing state-of-the-art approaches that incorporate resolution information via positional encoding. Implementation details for all baseline models are provided in the appendix.
\paragraph{Implementation Details}
Atomizer was trained for 40 epochs using a learning rate schedule consisting of a 5-epoch linear warmup followed by cosine annealing to zero. We used a batch size of 1024 distributed across two NVIDIA H100 GPUs. The model architecture consists of 4 cross-attention blocks, each containing 4 self-attention layers. To improve parameter efficiency, we employed weight sharing across all cross-attention blocks except the first. For classification output, we implemented latent attention pooling following \cite{chen2023self}. To reduce the computational overhead induced by the cross attention layer, we randomly apply pruning on 50\% of the model input tokens before each cross attention block.

\section{Results}

This section presents experimental results comparing Atomizer with baseline models (Perceiver \cite{jaegle2021perceiver}, ViT \cite{dosovitskiy2020image}, ResNet \cite{he2016deep}, and ScaleMAE \cite{reed2023scale}) across different settings, resolutions, and input sizes.

\begin{table}[ht]
    \centering
    \caption{Performance comparison of models across different test settings (mean Average Precision \%).}
    \label{tab:combined-results}
    \begin{tabular}{l|c|c|c|c|c|c}
        \toprule
        \textbf{Model} & \textbf{Test 1} & \textbf{Test 2} & \textbf{Test 3} & \textbf{Test 4} & \textbf{Test 5} & \textbf{Test 6} \\
        \midrule
        Perceiver   & 16.47 & 19.89 & 14.73 & 22.17 & 14.73 & 22.28 \\
        ViT         & 17.29 & 19.77 & 32.85 & 18.12 & 15.79 & 26.98 \\
        ResNet      & 17.91 & 19.37 & \underline{38.33} & 22.36 & 14.52 & 22.86 \\
        ScaleMAE    & \underline{20.89} & \underline{20.32} & 34.92 & \underline{25.70} & \underline{19.61} & 26.90 \\
        Atomizer    & \textbf{42.50} & \textbf{35.63} & \textbf{43.34} & \textbf{37.53} & \textbf{32.28} & \textbf{47.74} \\
        \bottomrule
    \end{tabular}
\end{table}

\noindent Table \ref{tab:combined-results} shows model performance across the two evaluation settings. On the standard BigEarthNet benchmark (with access to every band at 10m resolution), Atomizer achieves 48.66\% AP, outperforming all baseline models. The results on modality-disjoint tests (Test 1-6) provide further insights on generalization capability.
ViT and ScaleMAE perform similarly on BigEarthNet (33.47\% vs. 33.04\% AP), but exhibit different behavior on modality-disjoint tests. In general, ScaleMAE maintains higher performance on the test sets compared to ViT, indicating that its resolution-aware positional encoding benefits cross-modality generalization. This confirms the importance of incorporating resolution information, though ScaleMAE still underperforms compared to a fully modality-agnostic solution.
The Perceiver model's performance is particularly informative. Despite architectural similarities with Atomizer, both representing inputs as tokens mapped to a latent space via cross-attention, Perceiver achieves lower results across all settings (15.16\% AP on BigEarthNet). This difference demonstrates that the encoding scheme for tokens is a critical factor determining model performance on remote sensing data. Atomizer's approach to encoding spatial, spectral, and resolution information within tokens contributes substantially to its performance advantage.

\noindent In the following, we discuss model performance across different spatial resolutions (Table~\ref{tab:res-ap}) and input sizes  (Table~\ref{tab:size-ap}) on the original BigEarthNet. 
Table \ref{tab:res-ap} shows model performance across spatial resolutions from 20 to 80 m/px. Atomizer achieves 48.66\% AP at 20 m/px and 44.09\% AP at 80 m/px, representing a larger relative decrease than the rest of methods, but staying comfortably above them in all settings.
The nearly constant performance of Perceiver and ScaleMAE across resolutions aligns with its design goal of resolution invariance, but this comes at the cost of lower absolute performance compared to Atomizer.
These results indicate that Atomizer's token-based encoding of resolution as metadata enables effective processing across different spatial scales while maintaining higher performance than resolution-invariant approaches.

\begin{table}[h]
    \centering
    \caption{mean Average Precision (mAP) at different spatial resolutions (in meters per pixel). }
    \label{tab:res-ap}
    \begin{tabular}{lccccc}
        \toprule
        \textbf{Model} & \textbf{20 m/px} & \textbf{26.7 m/px} & \textbf{40 m/px} & \textbf{80 m/px} \\
        \midrule
        Perceiver   & 15.16 & 15.16 & 15.16 & 15.16 \\
        ViT         & \underline{33.47} & \underline{33.48} & 33.32 & 32.44 \\
        ResNet      & 29.50 & 32.50 & 32.37 & 30.57 \\
        ScaleMAE    & 33.04 & 33.08 & \underline{33.44} & \underline{33.14} \\
        Atomizer    & \textbf{48.66} & \textbf{48.56} & \textbf{47.88} & \textbf{44.09} \\
        \bottomrule
    \end{tabular}
\end{table}

\noindent Table \ref{tab:size-ap} presents model performance across input sizes from 30$\times$30 to 120$\times$120 pixels. Atomizer's performance increases with input size, from 37.98\% AP at 30$\times$30 pixels to 48.66\% AP at 120×120 pixels (a 28.1\% relative improvement). At all input sizes, Atomizer outperforms all models.
ResNet shows non-monotonic behavior, with performance peaking at 60$\times$60 pixels (31.30\% AP) before declining at 120$\times$120 pixels (29.50\% AP). ViT and ScaleMAE exhibit more consistent scaling patterns but with lower overall performance than Atomizer.
The Perceiver model maintains constant performance (15.16\% AP) regardless of input size. This suggests that despite its token-based design, Perceiver does not effectively utilize the additional spatial context in larger images. These results demonstrate the advantage of Atomizer's approach to encoding spatial positions within each token.

\begin{table}[h]
    \centering
    \caption{mean Average Precision (mAP) for varying input sizes (in pixels). }
    \label{tab:size-ap}
    \begin{tabular}{lcccc}
        \toprule
        \textbf{Model} & \textbf{30$\times$30 px} & \textbf{60$\times$60 px} & \textbf{90$\times$90 px} & \textbf{120$\times$120 px} \\
        \midrule
        Perceiver  & 15.161 & 15.161 & 15.161 & 15.161 \\
        ViT        & 23.734 & 30.408 & \underline{34.238} & \underline{33.474} \\
        ResNet     & \underline{26.858} & 31.297 & 26.339 & 29.495 \\
        ScaleMAE   & 26.112 & \underline{33.225} & 34.191 & 33.036 \\
        Atomizer   & \textbf{37.975} & \textbf{45.887} & \textbf{47.958} & \textbf{48.663} \\
        \bottomrule
    \end{tabular}
\end{table}

\section{Conclusions}

We have described Atomizer, a model that reasons on the atomic elements of an optical multi-spectral image: the individual scalar values that compose it. This allows to seamlessly modify the resolution, image size and spectral bands of the input images without any modification to the encoder.
Our experiments on cross-modal evaluation, where the resolution, size and bands during training time are not the same as at test time, suggest that the proposed approach provides substantially better generalization than all of the compared methods, all without access to the test modalities while training nor any type of adaptation procedure.
We believe these properties are fundamental for the building blocks of the next generation of Earth observation foundation models, which should be ready to process data coming from future satellite missions.
Although we have focused on single-time multi-spectral imagery derived from Sentinel-2, we will further investigate the extension of these principles to multi-temporal imagery and other types of sensors with different geometries, such as Synthetic Aperture Radar. In addition, there is a need for spatio-temporal output structures, allowing for tasks such as semantic segmentation and object detection.



\printbibliography[]
\appendix

\end{document}